\documentclass{article}

\usepackage{microtype}
\usepackage{graphicx}
\usepackage{booktabs} 
\usepackage{subcaption}

\usepackage{hyperref}

\usepackage[accepted]{icml2024}

\usepackage{amsmath}
\usepackage{amssymb}
\usepackage{mathtools}
\usepackage{amsthm}
\usepackage{wasysym}
\usepackage{upgreek}

\usepackage[capitalize,noabbrev]{cleveref}

\theoremstyle{plain}

\theoremstyle{definition}

\theoremstyle{remark}

\usepackage[textsize=tiny]{todonotes}

\begin{document}

\twocolumn[
\icmltitle{Predicting Stock Price Direction on Earnings Announcement Days using Multi-modal Deep Learning}

\vskip 0.3in

\begin{minipage}{\textwidth}
\centering
\begin{tabular}{ccc}
{\large Manuel Noseda} & {\large Nathan Soldati} & {\large Marco Paina} \\
ETH Z\"urich & ETH Z\"urich & ETH Z\"urich \\
\texttt{mnoseda@ethz.ch} & \texttt{nsoldati@ethz.ch} & \texttt{mpaina@ethz.ch} \\
\end{tabular}
\end{minipage}

\icmlkeywords{Deep Learning, NLP, Event Studies, Quantitative Finance}

\vskip 0.25in
]


\begin{abstract}
Predicting stock price movements during Earnings Announcements (EAs) is a significant challenge due to market noise and high-impact price discontinuities. In this study, we evaluate whether pre-announcement news sentiment, firm fundamentals, and recent market dynamics jointly predict the directional price movement of equities on EA days. We construct a multi-modal feature space combining 15 fundamental metrics, 3 price-based technical indicators and sentiment scores derived from financial news articles processed using FinBERT. We compare a Long Short-Term Memory (LSTM) network and a Transformer-based architecture against a logistic regression baseline, and further assess all models with and without sentiment features to quantify their incremental value. Our results indicate that while the LSTM demonstrates higher precision through a conservative safe-bet strategy, the Transformer model exhibits superior sensitivity in identifying volatile movements, achieving a higher macro F1-score, with ablation experiments showing a consistent benefit from incorporating news sentiment.
\end{abstract}

\section{Introduction}
Earnings Announcements (EAs) are associated with some of the most dramatic price changes in the stock market. Since the early work of \citet{BallBrown1968}, researchers have long known that accounting data can help predict how stock prices react to these events. Later, \citet{BernardThomas1989} showed that prices often continue to move for days after the announcement, a trend known as post-earnings announcement drift, indicating that the initial market reaction may be incomplete.

Recently, researchers have started looking beyond numbers. \citet{Tetlock2007} showed that news sentiment can capture information not present in financial reports. Early deep learning approaches explicitly targeting event-driven stock movements demonstrated that textual event representations can improve return prediction \citep{Ding2015}. With the rise of modern language models, architectures such as FinBERT \citep{Araci2019} allow us to measure this sentiment more accurately than ever before, as shown by recent work \citep{LiZhangZhao2023}.

However, most existing stock prediction models focus on day-to-day trading or very high-frequency movements. There is comparatively less research specifically focused on EAs, even though these moments are crucial for investors. Recent work has begun to address this gap by showing that the textual content of earnings press releases (“soft information”) can predict EA-day returns and is competitive with traditional “hard information” such as earnings surprises \citep{wu_et_al_2025}.

Our project differs from prior work in that we combine three different types of data: firm fundamentals, market prices and news sentiment from FinBERT. We compare a standard logistic regression baseline against LSTM and Transformer models, and perform an ablation study in which sentiment features are removed to assess their marginal contribution. The Transformer's attention mechanism is intended to identify which days in the 30-day period before an announcement are the most important. To handle the fact that stock prices are usually stable (neutral class dominates), we use a specialized loss function to focus on the rare, sharp moves (up/down classes). This approach aims to provide a more reliable tool for decision-making during these high-volatility events.

\section{Models and Methods}

\subsection{Problem Formulation}

We define our task as Event-Driven Price Movement Prediction. Given a multi-modal temporal sequence $\mathbf{X} = [\mathbf{x}_1, \mathbf{x}_2, ..., \mathbf{x}_T]$ representing states on the 30 days ($T=30$) prior to an Earnings Announcement (EA), the goal is to predict the price direction $y$ after the EA, where $y \in \{0 \text{ (UP)}, 1 \text{ (DOWN)},2\text{ (NEUTRAL)}\}$.

\subsection{Data Extraction}
\label{subsec:data_extraction}

The data extraction pipeline constructs the raw dataset by integrating financial news and firm-level fundamentals over a fixed ten-month window. News articles are sourced from a pre-downloaded Markets API feed spanning multiple periods and company batches, which are merged to recover the full firm cross-section and ensure continuous temporal coverage. Article timestamps are parsed, duplicates removed, and items sorted chronologically at the ticker level.

Firm fundamentals are retrieved from FactSet using a configuration-driven interface. The firm universe consists of the top 500 U.S. companies by market capitalization at the start of the sample period. For each firm, variables are queried and temporally aligned, and standard accounting ratios (e.g., margins, profitability, leverage) are computed following predefined specifications for data frequency, lags, and forward-filling.

All components are merged into a unified firm-indexed dataset containing price series, earnings dates, fundamental variables, derived ratios, and associated news articles. The resulting dataset is serialized and used as input for subsequent preprocessing, labeling, and modeling stages.

\subsection{Feature Engineering and Data Representation}

To capture the multi-faceted nature of equity markets, we construct a heterogeneous feature space with three modalities: firm fundamentals, market dynamics, and news sentiment.

For each ticker, we build a daily time series where each observation $\mathbf{x}_t \in \mathbb{R}^d$ represents the market state on day $t$. The final feature vector has dimensionality $d = d_f + d_m + d_n = 21$, composed of:
\begin{itemize}
    \item \textbf{Firm Fundamentals} ($d_f=15$): Quarterly and annual metrics such as Net Margin, ROE, Debt-to-Equity, and Operating Cash Flow, serving as proxies for financial health and intrinsic value.
     \item \textbf{Market Dynamics} ($d_m=3$): The adjusted closing price and two technical indicators (3-day and 6-day simple moving averages) capturing short-term momentum.
     \item \textbf{News Sentiment} ($d_n=3$): Financial news articles are processed using FinBERT, a language model fine-tuned for the financial domain. For each trading day, FinBERT produces a probability distribution over three sentiment classes (positive, negative, neutral). When multiple articles are published on the same day for a firm, their sentiment vectors are averaged to obtain a single daily representation. Textual information has been shown to contain predictive signals beyond standard firm characteristics \citep{KeKellyXiu2020}.
\end{itemize}

To assess the impact of sentiment, we also train all models using only fundamental and market-based features.

\subsection{Data Preprocessing and Imputation}
Financial data sparsity and scaling variances are addressed via a three-stage pipeline:
\begin{enumerate}
    \item \textbf{Temporal Imputation}: We utilize a Forward-Fill (FFILL) strategy for weekends and market holidays, assuming state stationarity until the next trading session.
    \item \textbf{Cross-Sectional Imputation}: Sparse fundamental metrics are handled via mean imputation to maintain feature density without introducing idiosyncratic bias.
    \item \textbf{Feature Scaling}: Finally, all features undergo $z$-score standardization ($\mu=0$, $\sigma=1$). This normalization is critical for the stability of gradient-based optimization in LSTMs and Transformers, preventing high-magnitude features (e.g., Total Assets) from dominating the loss function over smaller-scale features (e.g., Sentiment).
\end{enumerate}

\subsection{Target Definition and Imbalance}

The target variable $y$ is derived from the security’s adjusted closing price on the earnings announcement day (or the following trading day if the announcement occurs after market close), denoted $P_{EA}$, relative to the preceding trading day price $P_{EA-1}$. To filter short-term noise and isolate significant volatility, we define the daily return $R = (P_{EA} - P_{EA-1}) / P_{EA-1}$ and apply a classification threshold $\tau = 3\%$, yielding the three-class target $y \in \{0,1,2\}$ as defined in \eqref{eq:target_definition}.

\begin{equation}
\label{eq:target_definition}
y = \begin{cases} 
0 \text{ (UP)}, & \text{if } R \geq \tau \\ 
1 \text{ (DOWN)}, & \text{if } R \leq -\tau \\ 
2 \text{ (NEUTRAL)}, & \text{otherwise} 
\end{cases}
\end{equation}

The resulting labels are highly imbalanced, with the \textsc{NEUTRAL} class accounting for approximately 67\% of observations. To mitigate this, we employ a \textbf{weighted cross-entropy loss} that assigns higher weights \(w_k\) to minority classes, penalizing costly directional errors (e.g., \textsc{DOWN} vs.\ \textsc{UP}) more strongly. The loss is defined as \(\mathcal{L} = - \sum_{k=0}^{2} w_k\, y_k \log(\hat{y}_k)\), with weights chosen such that \(w_0, w_1 \geq w_2\).

\subsection{Model Architectures}

We evaluate two deep learning models for capturing temporal dependencies in the 21-dimensional feature space, a recurrent LSTM and a self-attention-based Transformer, and compare them to a logistic regression baseline.

\subsubsection*{Baseline}
As a non-temporal baseline, we use a multinomial logistic regression classifier. Each $30$-day input window is flattened by concatenating daily observations into a fixed-length vector of size $30 \times d$. The model is trained with the \texttt{lbfgs} solver and class-balanced weights to partially address label imbalance, providing a linear reference for assessing the benefit of explicit temporal modeling.

\subsubsection*{LSTM}
We implement a $2$-layer stacked LSTM with hidden dimension $64$ and dropout rate $0.5$. Daily feature vectors $\mathbf{x}_t \in \mathbb{R}^d$ are processed sequentially, and the final hidden state is passed to a fully connected softmax layer to produce class probabilities.

\subsubsection*{Transformer with Self-Attention}

To relax the inductive bias of strictly sequential processing, we implement a Transformer encoder architecture. Following observations by \citet{LiZhangZhao2023}, we evaluate self-attention not as a guaranteed improvement over recurrent models, but for its ability to capture global temporal patterns.

The model uses multi-head self-attention to capture dependencies across the entire $30$-day pre-announcement window. Temporal order is preserved via sinusoidal positional encodings added to the input embeddings.

Our configuration consists of $2$ encoder layers with $4$ attention heads and a feed-forward dimension of $256$. After the final encoder block, global average pooling is applied over the temporal dimension, followed by a fully connected classification head. This architecture enables joint modeling of fundamental, market, and sentiment features while remaining compact and suitable for limited data.

\subsection{Experimental Setup}

The models were implemented in PyTorch and trained on the ETH Student Cluster. All models were optimized using the Adam optimizer with a learning rate of \(5 \cdot 10^{-5}\) and a batch size of \(8\). Regularization was applied through a fixed training duration of \(15\) epochs and a dropout rate of \(0.5\) for both the LSTM and Transformer architectures.

Model performance is evaluated using metrics aligned with financial decision-making rather than raw accuracy. Since the majority class (\textsc{NEUTRAL}) corresponds to a \textit{no-action} state, directional errors are substantially more costly than missed opportunities. We therefore report \textbf{Precision} to assess the reliability of actionable (\textsc{UP}/\textsc{DOWN}) signals, and the \textbf{Macro-averaged F1-score} to ensure balanced performance across imbalanced classes. Macro-F1 is computed by first evaluating the F1-score independently for each class and then averaging these scores across all classes, giving equal importance to \textsc{UP}, \textsc{DOWN}, and \textsc{NEUTRAL} outcomes regardless of their frequency.

In addition, we introduce a \textbf{custom cost metric} to explicitly encode asymmetric financial risk. Correct predictions incur zero cost; \textsc{UP}$\leftrightarrow$\textsc{DOWN} misclassifications incur a cost of $3$, while all other errors incur a cost of $1$. The cost is averaged across samples, yielding an expected per-observation loss that complements Macro-F1 by reflecting domain-specific risk preferences. It is not optimized directly, but used as an evaluation proxy.

To ensure reproducibility, the complete implementation, including preprocessing scripts, model architectures, and training logs, is available at our GitHub repository (\url{https://github.com/Marc0-Pol0/dl-project}).

\section{Results}

\subsection{Confusion Matrices and Error Analysis}

\begin{figure}[!ht]
    \centering
    \begin{subfigure}[t]{0.49\linewidth}
        \centering
        \includegraphics[width=\linewidth]{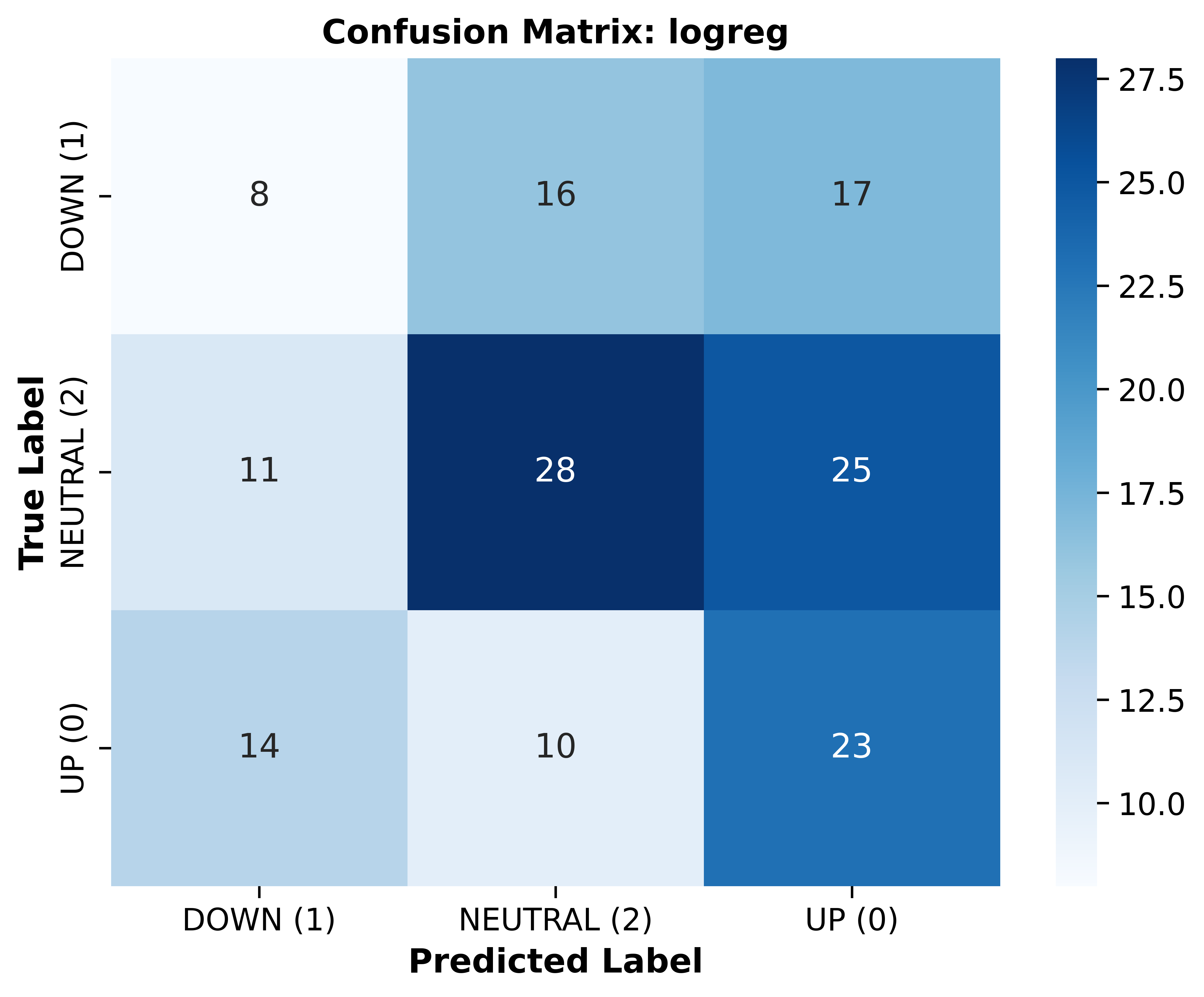}
        \caption{Baseline}
        \label{fig:cf_baseline}
    \end{subfigure}
    \hfill
    \begin{subfigure}[t]{0.49\linewidth}
        \centering
        \includegraphics[width=\linewidth]{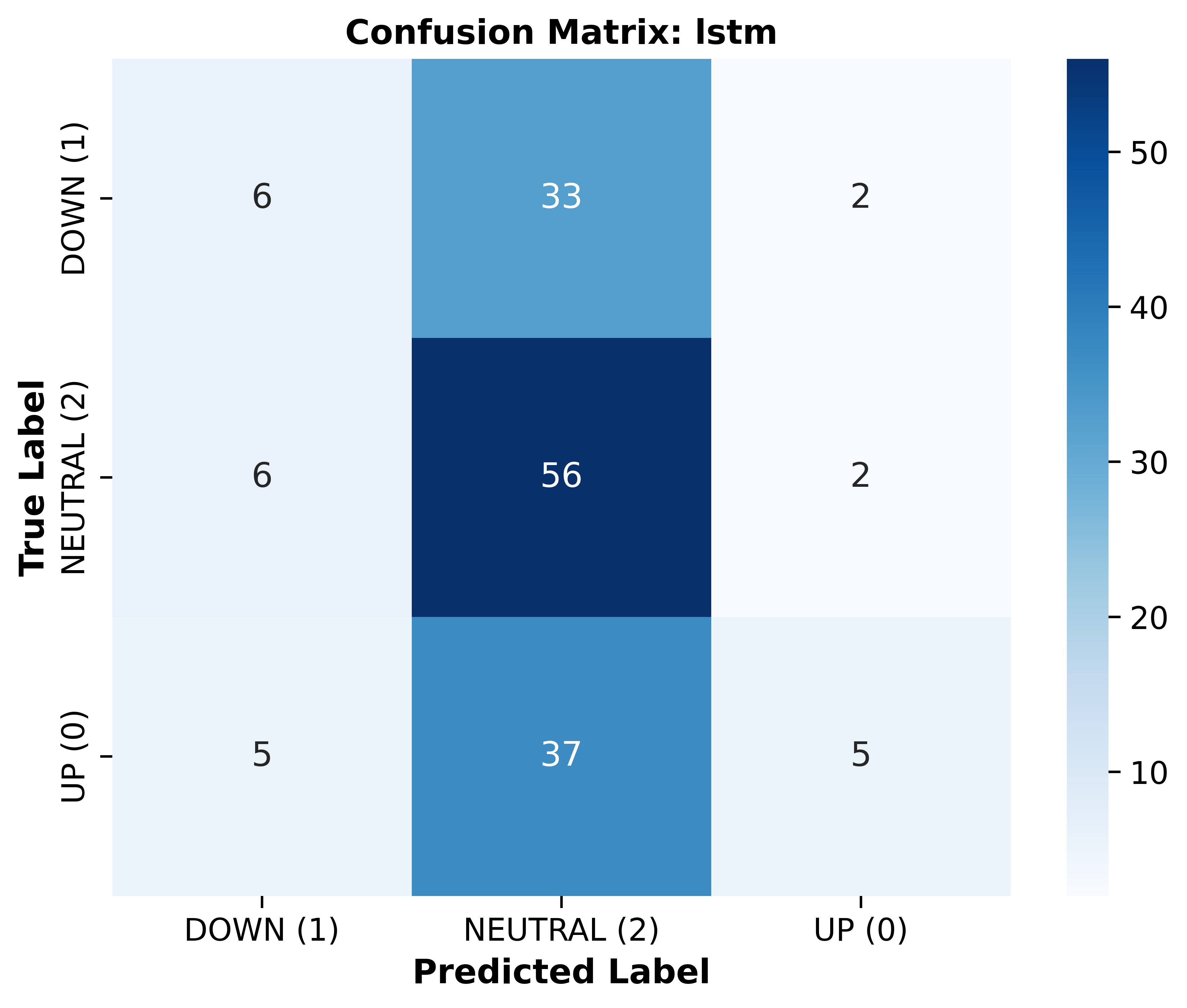}
        \caption{LSTM}
        \label{fig:cf_lstm}
    \end{subfigure}

    \vspace{2mm}

    \begin{subfigure}[t]{0.5\linewidth}
        \centering
        \includegraphics[width=\linewidth]{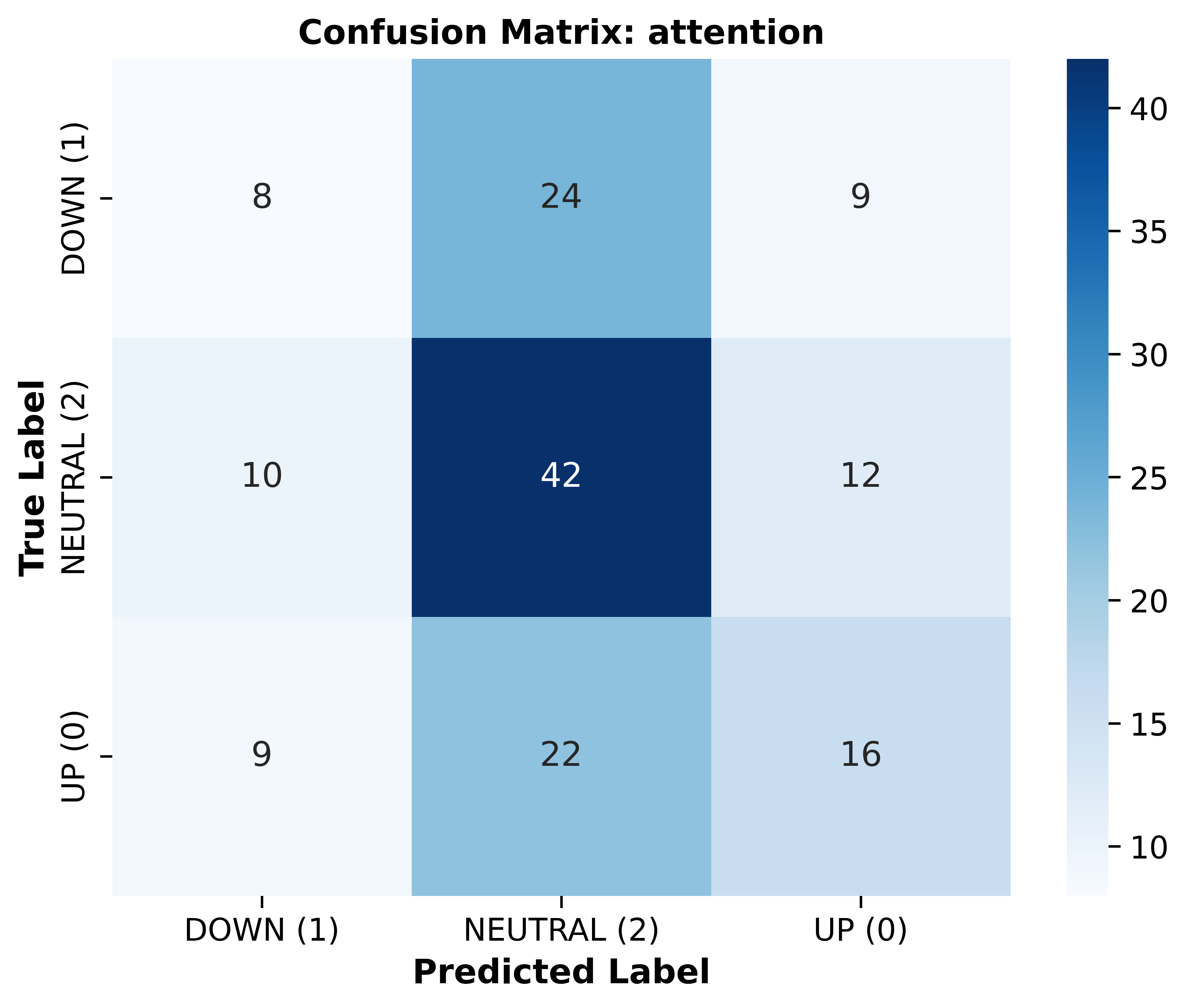}
        \caption{Attention}
        \label{fig:cf_attention}
    \end{subfigure}

    \caption{Confusion matrices for the evaluated models.}
    \label{fig:confusion_matrices}
\end{figure}

Figure~\ref{fig:confusion_matrices} presents the confusion matrices for the three models trained with sentiment features.
For completeness, larger confusion matrices for all six model variants (with and without sentiment) are reported in Appendix~\ref{app:confusion_matrices}.

The logistic regression baseline shows an unstable decision profile, with a strong bias toward predicting \textsc{UP} movements. This results in a large number of costly \textsc{UP}$\leftrightarrow$\textsc{DOWN} swaps, limiting its practical usefulness despite a Macro-F1 comparable to the deep models.

The LSTM adopts a highly conservative strategy, predicting \textsc{NEUTRAL} in most cases. This substantially reduces directional swaps and leads to the lowest custom cost, but also limits sensitivity to volatile events, yielding a lower Macro-F1 score.

The Attention model distributes predictions more evenly across classes. It identifies more \textsc{UP} and \textsc{DOWN} events than the LSTM, improving sensitivity, but at the expense of increased directional errors and higher financial risk.

\subsection{Performance Comparison}

Table~\ref{tab:results} summarizes performance across all models. The Attention model achieves the highest Macro-F1 score, reflecting superior sensitivity to earnings-related volatility. In contrast, the LSTM minimizes financial risk, attaining the lowest custom cost through conservative predictions. Logistic regression consistently incurs the highest cost due to frequent directional misclassifications.

Removing sentiment features degrades performance for all models, increasing custom cost and reducing Macro-F1. This effect is most pronounced for the LSTM, indicating that sentiment information plays a stabilizing role in deep sequential models.

\begin{table}[h!]
    \centering
    \scriptsize
    \setlength{\tabcolsep}{2pt}
    \begin{tabular}{l|c|cccccc}
        \toprule
        Model & Sent. & Acc. (\%) & Macro-F1 & Prec. (UP) & Prec. (DOWN) & Custom Cost\\
        \midrule
        Log reg   & \checked & 38.816 & 0.367 & 0.354 & 0.242 & 1.020 \\
        LSTM      & \checked & \textbf{44.079} & 0.325 & \textbf{0.556} & \textbf{0.353} & \textbf{0.651} \\
        Attention & \checked & 43.421 & \textbf{0.390} & 0.432 & 0.296 & 0.803 \\
        \midrule
        Log reg   & $\upchi$  & 37.500 & 0.370 & 0.339 & 0.240 & 1.138 \\
        LSTM      & $\upchi$  & 38.816 & 0.292 & 0.000 & 0.274 & 0.914 \\
        Attention & $\upchi$  & 39.474 & 0.359 & 0.385 & 0.261 & 0.921 \\
        \bottomrule
    \end{tabular}
    \caption{Model Performance Comparison.}
    \label{tab:results}
\end{table}

Overall, the results reveal a clear trade-off between sensitivity and risk control. The LSTM prioritizes stability and cost minimization, while self-attention improves detection of volatile price movements at the expense of higher directional risk.

\section{Discussion}

Our results highlight a clear trade-off between conservative and sensitivity-oriented modeling strategies for earnings announcement prediction. The LSTM exhibits a strong bias toward the \textsc{NEUTRAL} class, resulting in higher precision and fewer critical directional errors, but at the cost of missing a large fraction of volatile events. In contrast, the Transformer-based model is more responsive to pre-announcement signals and captures more price movements, albeit with a higher false positive rate and increased directional risk.

The logistic regression baseline lacks a stable decision strategy, producing frequent high-cost directional errors despite achieving a Macro-F1 comparable to the deep models. This confirms that linear models are insufficient for capturing the complex temporal and cross-modal interactions present in earnings-related data.

The ablation study further shows that removing sentiment features degrades performance across all models, particularly increasing directional errors. This indicates that news sentiment provides complementary information that stabilizes predictions beyond fundamentals and price dynamics alone. The fact that removing sentiment degrades performance is consistent with empirical evidence that aggregated news sentiment contains predictive information about stock returns beyond fundamentals and prices, particularly when considered over longer horizons and around earnings events \citep{heston_sinha_2016}.

Overall, model choice should depend on the target application: risk-averse strategies may favor the LSTM, while volatility-seeking strategies may benefit from the Transformer's higher sensitivity. Both approaches, however, remain constrained by limited data and the intrinsic noise of financial markets.

\section{Summary}

In this work, we studied the problem of predicting directional stock price movements on earnings announcement days using a multi-modal feature set combining firm fundamentals, market dynamics, and news-based sentiment. We compared a logistic regression baseline with LSTM and Transformer architectures designed to model temporal dependencies in the pre-announcement window.

Our results show that the LSTM achieves lower financial risk through conservative predictions, while the Transformer captures more volatile movements and attains a higher Macro-F1 score. The baseline performs poorly in terms of risk control, and sentiment features consistently improve stability and performance across models. These findings underscore the inherent trade-offs between sensitivity and risk management when applying deep learning models to noisy, imbalanced financial prediction tasks.

\section*{Acknowledgments}
We thank the teaching team of the ETH Z\"urich Deep Learning course (HS2025) for their guidance and feedback throughout the project.

\newpage
\onecolumn
\bibliographystyle{icml2024} 
\bibliography{example_paper}


\appendix
\section{Extended Confusion Matrices}
\label{app:confusion_matrices}

\begin{figure}[ht]
    \centering
    \begin{subfigure}[t]{0.49\linewidth}
        \centering
        \includegraphics[width=0.99\linewidth]{cm_logreg.png}
        \caption{With sentiment}
    \end{subfigure}
    \hfill
    \begin{subfigure}[t]{0.49\linewidth}
        \centering
        \includegraphics[width=0.99\linewidth]{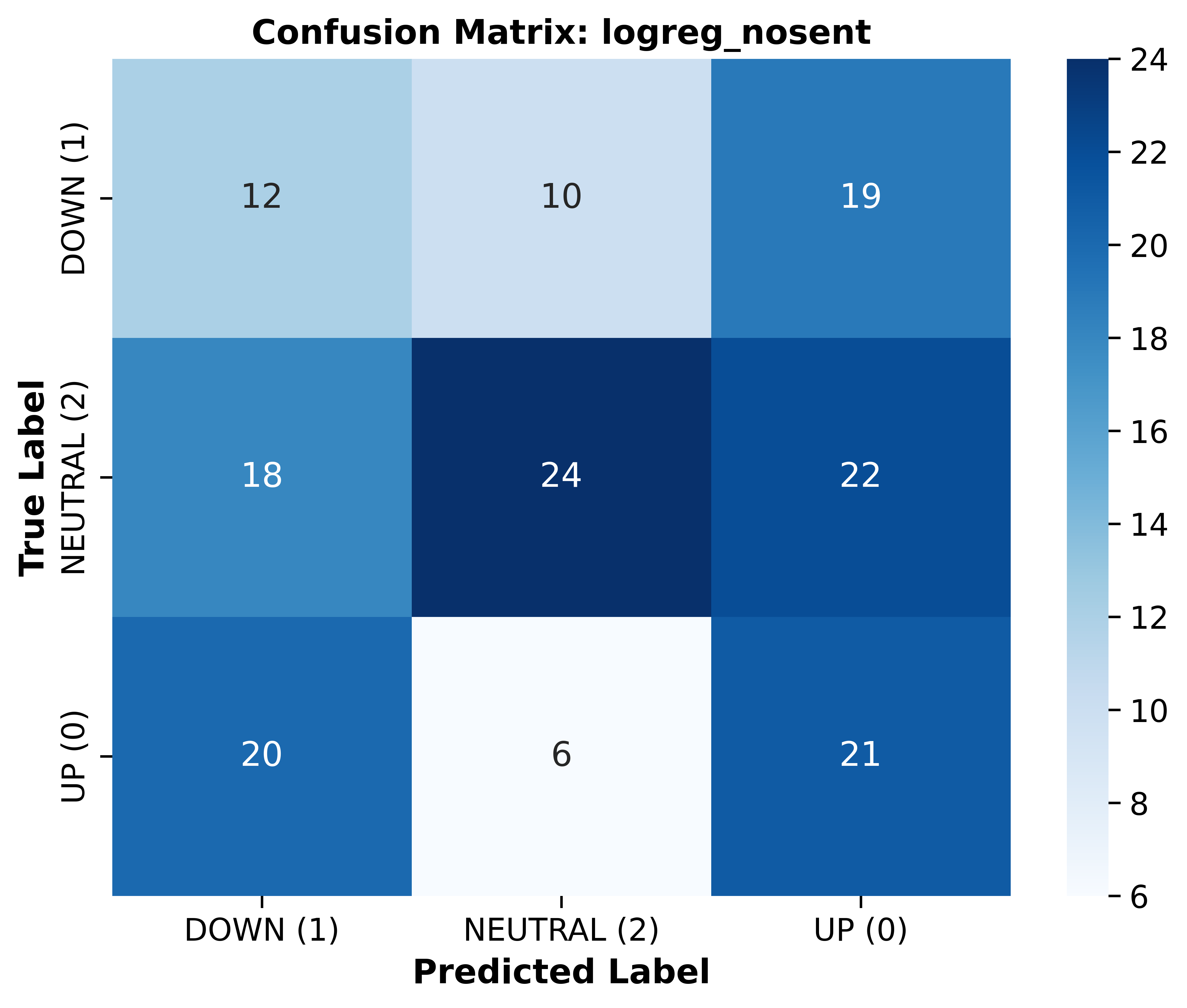}
        \caption{Without sentiment}
    \end{subfigure}
    \caption{Logistic Regression confusion matrices.}
    \label{fig:cm_logreg_appendix}
\end{figure}

\begin{figure}[ht]
    \centering
    \begin{subfigure}[t]{0.49\linewidth}
        \centering
        \includegraphics[width=0.9\linewidth]{cm_lstm.png}
        \caption{With sentiment}
    \end{subfigure}
    \hfill
    \begin{subfigure}[t]{0.49\linewidth}
        \centering
        \includegraphics[width=0.99\linewidth]{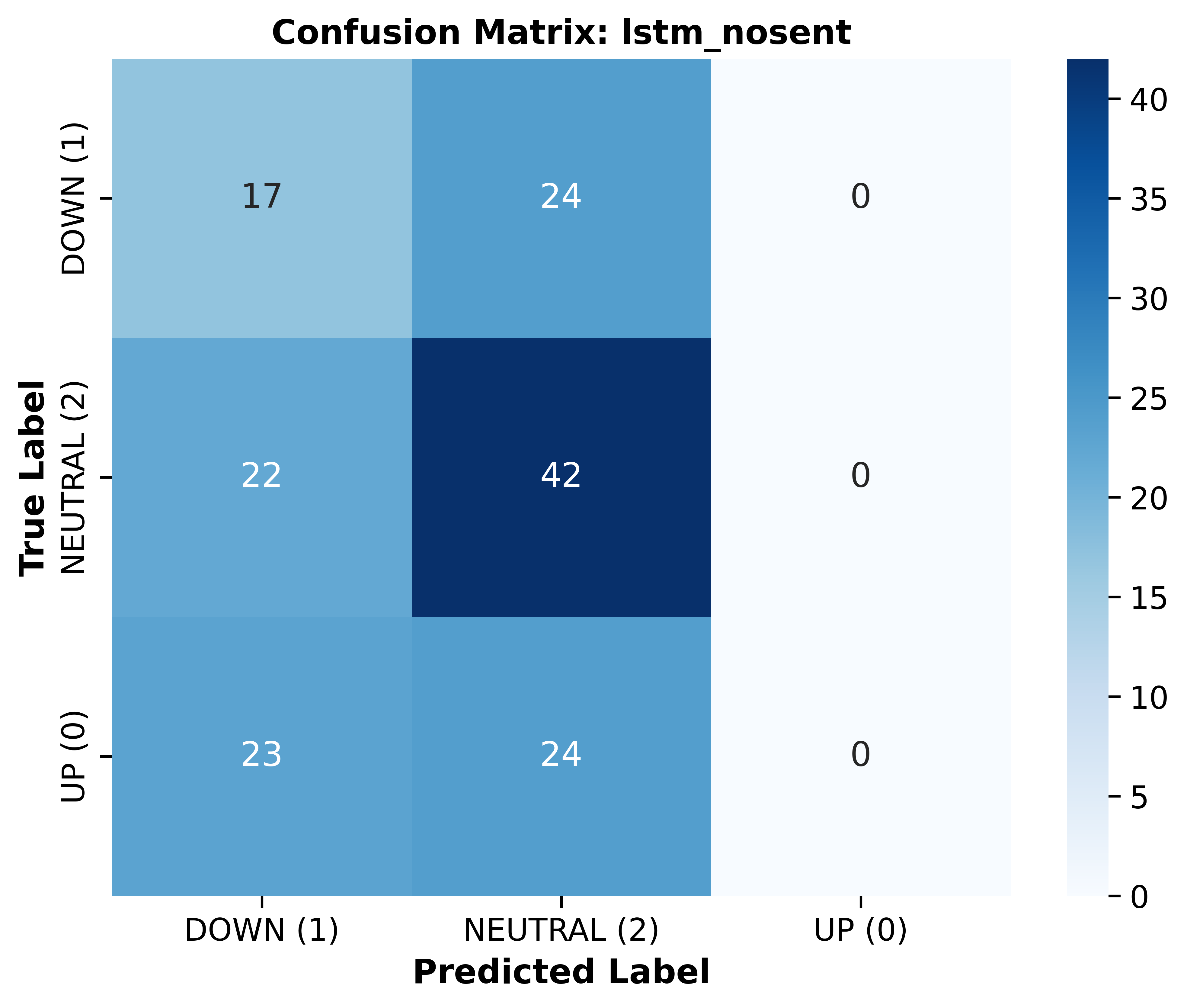}
        \caption{Without sentiment}
    \end{subfigure}
    \caption{LSTM confusion matrices.}
    \label{fig:cm_lstm_appendix}
\end{figure}

\begin{figure}[ht]
    \centering
    \begin{subfigure}[t]{0.49\linewidth}
        \centering
        \includegraphics[width=0.9\linewidth]{cm_attention.png}
        \caption{With sentiment}
    \end{subfigure}
    \hfill
    \begin{subfigure}[t]{0.49\linewidth}
        \centering
        \includegraphics[width=0.9\linewidth]{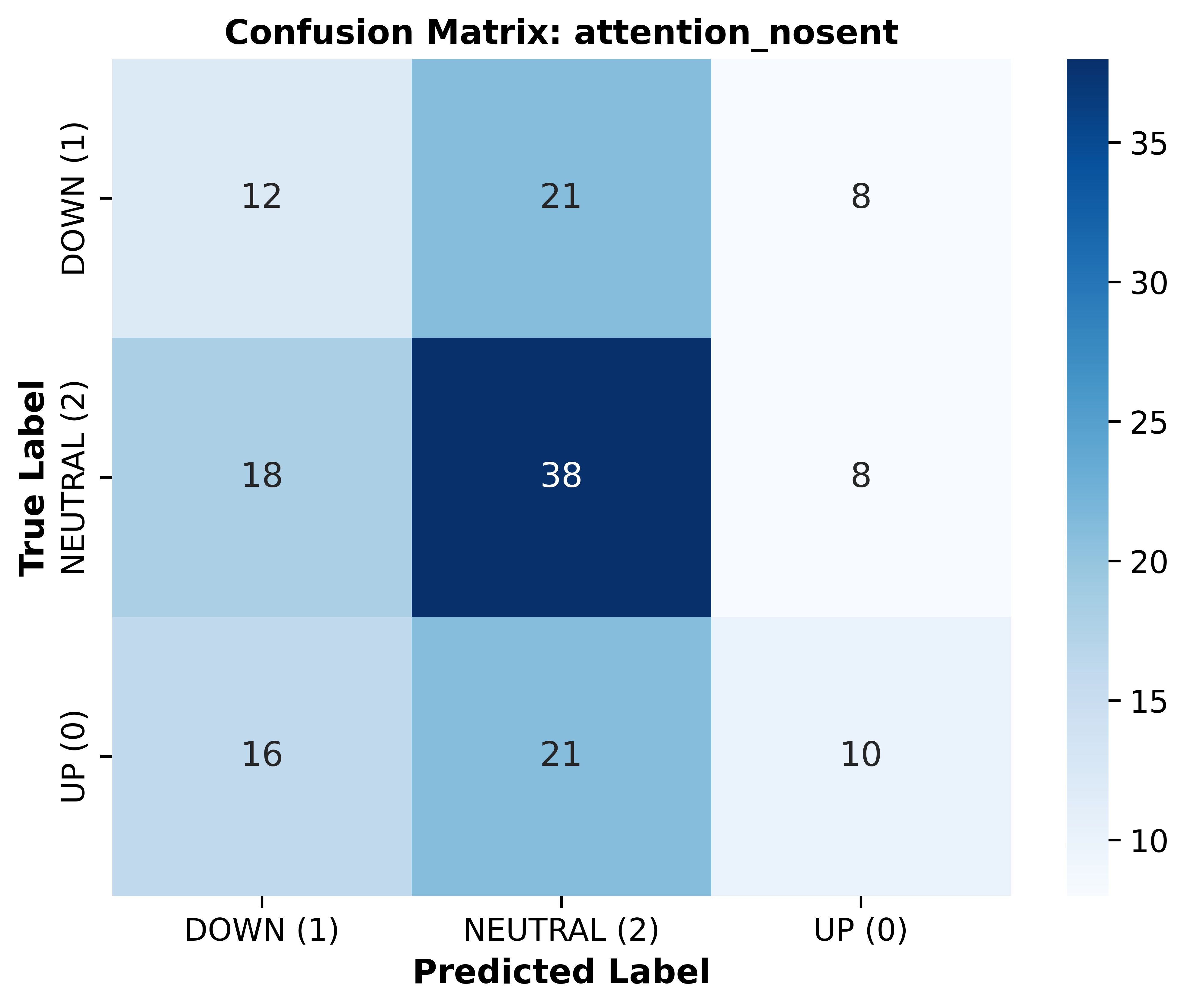}
        \caption{Without sentiment}
    \end{subfigure}
    \caption{Transformer confusion matrices.}
    \label{fig:cm_attention_appendix}
\end{figure}

\end{document}